\documentclass{bmvc2k}
\usepackage{multirow}
\usepackage[section]{placeins}
\usepackage{amssymb}
\pdfoutput=1
\usepackage{color}
\title{Self-Supervised Monocular Depth Estimation with Internal Feature Fusion}

\addauthor{Hang Zhou}{hang.zhou@uea.ac.uk}{1}
\addauthor{David Greenwood}{david.greenwood@uea.ac.uk}{1}
\addauthor{Sarah Taylor}{s.l.taylor@uea.ac.uk}{1}

\addinstitution{
 School of Computing Sciences
 University of East Anglia\\
 Norwich, UK
}

\runninghead{Zhou at el}{DIFFNet for self-supervised Monocular Depth Estimation}

\def\etal{\emph{et al}\bmvaOneDot}

\begin{document}
\maketitle
\begin{abstract}
 Self-supervised learning for depth estimation uses geometry in image sequences for supervision and shows promising results. Like many computer vision tasks, depth network performance is determined by the capability to learn accurate spatial and semantic representations from images. Therefore, it is natural to exploit semantic segmentation networks for depth estimation. 
%  In this work, based on a well-developed semantic segmentation network HRNet, we propose a novel \textbf{D}epth estimation network with \textbf{I}nternal \textbf{F}eature \textbf{F}usion, \textbf{DIFFNet}.
 In this work, based on a well-developed semantic segmentation network HRNet, we propose a novel depth estimation network \textbf{DIFFNet}, which can make use of semantic information in down and up sampling procedures. 
 By applying feature fusion and an attention mechanism, our proposed method outperforms the state-of-the-art monocular depth estimation methods on the KITTI benchmark. Our method also demonstrates greater potential on higher resolution training data. We propose an additional extended evaluation strategy by establishing a test set of challenging cases, empirically derived from the standard benchmark. The code and trained models are available at \url{https://github.com/brandleyzhou/DIFFNet}. 
\end{abstract}
%
%-------------------------------------------------------------------------
%
\section{Introduction}
\label{sec:intro}
Monocular depth estimation methods predict the depth of a scene from a single image. As an upstream task for scene understanding, it has a wide range of practical applications including autonomous vehicles, robotics and 3D reconstruction. While specialist hardware such as LiDAR or RGB-D cameras can be employed in such applications, deriving 3D geometry from a monocular RGB camera remains compelling. Supervised depth estimation methods~\cite{Eigen2014,Miangoleh2021Boosting,chen2021s,yang2019inferring,GargDualPixelsICCV2019,Ranftl2020} can produce dense depth maps but require large amounts of labelled data, which can be costly and time consuming even with the help of depth sensors. Self-supervised methods~\cite{zhou_sfmlearner, monodepth2, chen2019towards, zhou2020constant,klingner2020self} only rely on the scene's geometry in sequential images, and can take advantage of large scale unlabelled training resources to gain an advantage over supervised approaches.

As a result of using Structure from Motion (SfM) to construct the supervisory signal, most self-supervised methods suffer from those pixels which violate the assumptions of SfM, for example low-texture, occlusion and moving objects. To alleviate this intrinsic problem, prior works seek further auxiliary constraints such as optical flow~\cite{chen2020} and semantics~\cite{kendall2018multi, Zhu_2020_CVPR} to collaborate with geometry information. In contrast with optical flow, semantic segmentation has a closer relationship to depth estimation. Intuitively, when human vision systems estimate scene depth, extracting semantic information is a critical part of this procedure; objects belonging to different categories have corresponding cues in depth perception. Visually, outputs from semantic networks and depth networks both need accurate object boundaries, which mainly determine the performance of a model~\cite{lyu2020hr}.

With the goal of improving depth estimation by semantic segmentation, most related works \cite{chen2019towards, kumar2021syndistnet, klingner2020self, choi2020safenet} require a separate well-trained semantic network either to guide representation learning in depth networks or generate masks to filter those pixels belonging to non-rigid objects. When training a self-supervised depth network with an extra supervised semantic network that requires ground truth labels, the most attractive advantage of self-supervised learning disappears, and other problems are introduced such as domain gap.

When we look inside depth networks~\cite{zhou_sfmlearner, monodepth2}, we find that all of them are based on an encoder-decoder architecture~\cite{u-net}, which uses skip connections to restore semantic and spatial information.
We propose a new representation learning network, \textbf{DIFFNet}, to explicitly utilize built-in semantic information effectively, based on a well-developed semantic network~\cite{hrnet2020}.
Our contributions are:
\textbf{(1)} We apply a novel internal feature fusion mechanism to a semantic network for depth estimation, to bridge the semantic gap between encoder and decoder feature maps.
\textbf{(2)} We propose an effective attention module in the decoder to process skip connections. 
%\textbf{(3)} We propose a new evaluation framework where approaches can be better evaluated using the difficult cases in the benchmark data. The difficult benchmark is built in a self-established manner.
\textbf{(3)} Our proposed method advances the state-of-the-art on the KITTI benchmark and outperforms other methods on a customised benchmark.
\textbf{(4)} We propose an extended evaluation strategy where methods can be further tested using difficult cases in the benchmark data, formed in a self-established manner.
% 
%-------------------------------------------------------------------------
\section{Related Work}
\label{related_work}
% In this section, we review the literature on methods for self-supervised depth estimation and representation learning.
% 
\subsection{Self-supervised Monocular Depth Estimation}
Inferring depth from a single image is an ill-posed problem as 3D points from multiple depth planes can be projected onto the same 2D pixel of an image. Inspired by a classic computer vision algorithm SfM, the seminal work of~\cite{zhou_sfmlearner} proposed a fundamental framework consisting of a depth network and a pose network which are trained simultaneously with sequential video frames. Many works~\cite{monodepth2, watson2021temporal, poggi2020uncertainty, wang2020self, kuznietsov2021comoda, gonzalezbello2020forget,johnston2020self, guizilini2020, zhou2020constant} have further developed this idea in terms of the objective functions or model architectures. Monocular depth estimation is now one of the most successful applications of self-supervised learning, and even outperforms supervised methods.  
\subsection{Semantic Information and Depth Estimation}
Semantic information has been introduced as an additional source for improving depth estimation. Prior works can be divided into two categories. The first uses a separate semantic segmentation model to either add constraints to a photometric loss or to distinguish pixels belonging to categories which violate the static-world assumption (e.g. pedestrians, moving vehicles).
A schema to deal with moving dynamic-class objects to avoid contamination to the photometric loss is found in \cite{klingner2020self}.
Motivated by the observation that semantic segmentation networks trained with limited ground truth can generate more defined object borders than that of depth estimation, Zhu \etal \cite{Zhu_2020_CVPR} proposed a measurement of border consistency between segmentation and depth, and minimized it to push a depth network towards more accurate edges. 
The second category of models to exploit semantic information contain those that use it for representation learning, rather than in the photometric loss. Chen \etal \cite{chen2019towards} measured the content consistency between depth and semantic maps to propose an additional supervisory signal which guides networks to learn semantic-rich features. Several prior works  \cite{guizilini2020semantically, choi2020safenet,kumar2021syndistnet} used a pretrained semantic network to guide the feature extraction of a depth network. Generally, most methods in this direction require an extra semantic network trained with labeled data. 
\subsection{Representation Learning for Monocular Depth Estimation}
 For extracting features from the input images~\cite{zhou_sfmlearner} proposed DispNet which was based on U-Net~\cite{Unet}, a typical encoder-decoder architecture.
 Monodepth2 \cite{monodepth2} proposed a feature encoder based on ResNet~\cite{he2016deep} which has since become the standard approach.
%  Then with the help of Resnet~\cite{he2016deep}, \cite{monodepth2} proposes a Resnet-based feature extractor which have become a mainstream for late-comers.
To increase the robustness of the photometric loss, \cite{shu2020feature} used an external network to transform a reference frame and target frames into another domain in which there are better alternative representations for texture-less regions. Guizilini \etal \cite{guizilini2020} introduced 3D convolutions to construct packing and unpacking blocks, which are the replacement of standard downsample and upsample operations and preserve more details in feature maps than those of 2D convolutions.
 
To bridge the semantic gap between the encoder and decoder in the depth network, ~\cite{lyu2020hr} redesigned the skip connections in a U-Net architecture by fusing features at different scales.
Kendall \etal \cite{kendall2018multi} proposed a multi-task training framework in which geometry and semantic representations are learned with a shared encoder.
Under this schema, their models for depth estimation, semantic segmentation and instance segmentation all outperform the competitors which were trained individually on each task. 
Inspired by these ideas, we investigate a network architecture which has two key attributes: an architecture suitable for semantic segmentation and depth estimation and an internal mechanism which evolves multiple chances for feature fusion. Therefore, we choose HRNet~\cite{hrnet2020} as our new encoder blueprint. HRNet is able to learn high-resolution representations from images that are both semantically and spatially descriptive, and has been successfully applied to human pose estimation, semantic segmentation and object detection.  
%-----------------------------------------------------------------------

\section{Self-supervised Monocular Depth Estimation Framework}
Our general framework is based on the SfM paradigm that is followed by all other self-supervised monocular depth estimation approaches. It requires a depth model $\Theta_{\text{depth}}$ and a pose model $\Theta_{\text{pose}}$ trained simultaneously with a triplet of sequential RGB frames \(I_t \in \mathbb{R}^{ H \times W \times 3 },  t \in \{-1, 0, 1\} \). At training time $\Theta_{\text{depth}}$ takes a target frame $I_{0}$ as input and predicts a depth map $d = \Theta_{\text{depth}}(I_{0})$, while a relative pose change between the target frame and a source frame is estimated, \(T_{0 \rightarrow t^{'}} = \Theta_{\text{pose}}(I_{0},I_{t^{'}}), t^{'} \in \{-1,1\}\). 

Based on the assumption that the world is static and the view change is only caused by a moving camera, a synthesized counterpart to target frame $I_{0}$ can be generated using only pixels from the source frames \(I_{t^{'}}, t^{'} \in \{-1,1\} \):
\begin{equation}\label{eq:image_warp}
    I_{t^{'} \rightarrow 0 } = I_{t^{'}}[proj(reproj(I_{0},d,T_{0 \rightarrow t^{'}}),K)]
\end{equation}
where $K$ are known camera intrinsics, [] is the sampling operator, $reproj$ returns a 3D point cloud of camera $t^{'}$, and $proj$ outputs the 2D coordinates when projecting the point cloud onto $I_{t^{'}}$.     
Using the predicted depth map $d$, the generated view $I_{t^{'} \rightarrow 0 }$ and the corresponding target frame $I_{0}$, we build a supervisory signal consisting of two items:  

\noindent\textbf{Photometric Loss}, $ \ell_{p} $, is an appearance matching loss which calculates the difference between $I_{0}$ and $I_{t^{'} \rightarrow 0 }$. Following~\cite{monodepth2,Godard17}, the similarity between a synthesized frame and a target frame is computed using a Structural Similarity term (SSIM)~\cite{wang2004}. Then combining with the L1 norm, the final photometric loss function is defined: 
\begin{equation}\label{eq:photometrciloss}
    \ell_{p}(I_{0}, I_{t^{'} \rightarrow 0}) = \alpha \frac{1-SSIM(I_{0},I_{t^{'} \rightarrow 0})}{2} + (1-\alpha)|I_{0} - I_{t^{'} \rightarrow 0}| 
\end{equation}

\noindent\textbf{Edge-aware Smoothness}~\cite{Godard17}, $ \ell_{s}$, regularizes the depth in low gradient regions:
\begin{equation}\label{eq:smoothness}
    \ell_{s}(d) = |\frac{\nabla d}{\partial x}|e^{-|\frac{\nabla I_{0}}{\partial x}|} + |\frac{\nabla d}{\partial y} |e^{-|\frac{\nabla I_{0}}{\partial y}|}
\end{equation}
We also employ the minimum photometric error, auto-masking and multi-scale depth loss techniques which were introduced in~\cite{monodepth2}. The final self-supervised loss function is defined:
\begin{equation}\label{eq:lossfunction}
    \ell_{final} = min( \ell_{p}(I_{0}, I_{t^{'} \rightarrow 0}) ) + \beta\ell_{s}(d), t^{'} \in \{-1,1\}
\end{equation}
Where $\beta$ is a weighting coefficient between the photometric loss $\ell_{p}$ and depth smoothness $\ell_{s}$. The objective loss is averaged per pixel, pyramid scale and image batch.

%-------------------------------------------------------------------------
\section{DIFFNet}\label{sec:diffnet}
\begin{figure*}[t]
\centering
\includegraphics[width=\linewidth]{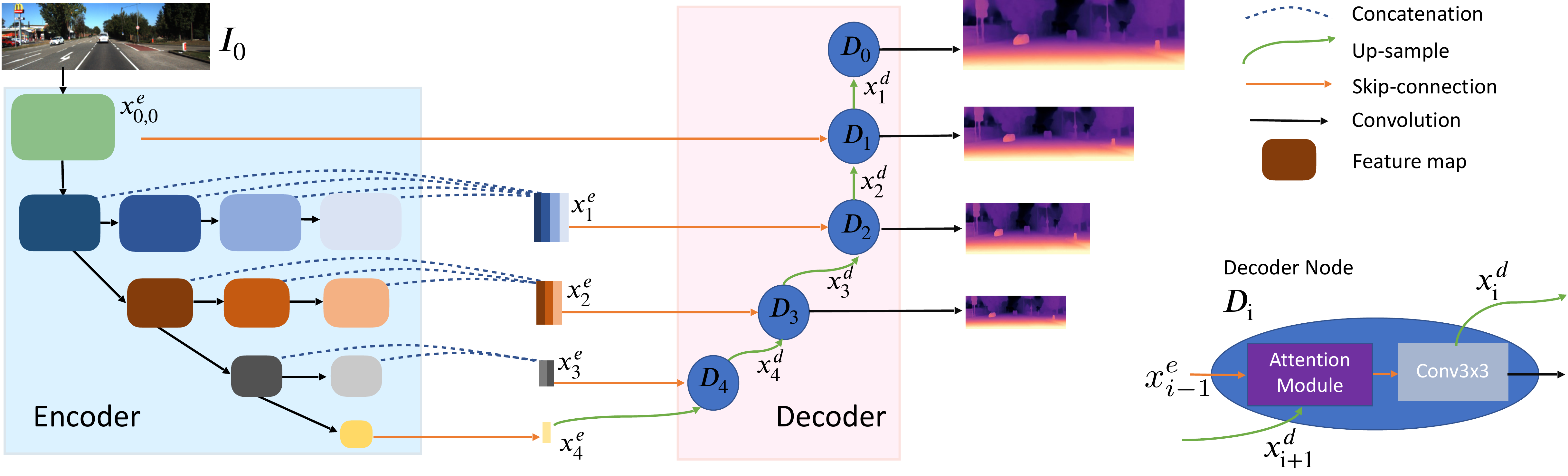}
\caption{An overview of the DIFFNet depth network. The encoder uses feature fusion to generate stacks of multi-stage feature maps. The decoder uses an attention module and a $3 \times 3 $ convolution layer to restore compressed feature maps at different scales.}
\label{fig:overview}
\end{figure*}

DIFFNet introduces a novel depth network which combines multiple resolution feature fusion and a spatial attention mechanism. In this section we provide details on our proposed network, which is built on an encoder-decoder architecture and is illustrated in Figure~\ref{fig:overview}.

%-------------------------------------------------------------------------
% 
\subsection{High-Resolution Depth Encoder}\label{sec:encoder}
Low level but high resolution features are spatially precise, and, conversely, high level but low resolution features are not spatially precise but are semantically rich. Many existing depth estimation approaches~\cite{monodepth2} are built on ResNet which encodes the input image as a low-resolution feature map. Instead, we investigate an effective architecture that is capable of fusing semantically-rich and spatially-precise features.

\begin{figure}
\begin{tabular}{cc}
\bmvaHangBox{\includegraphics[height=2.4cm]{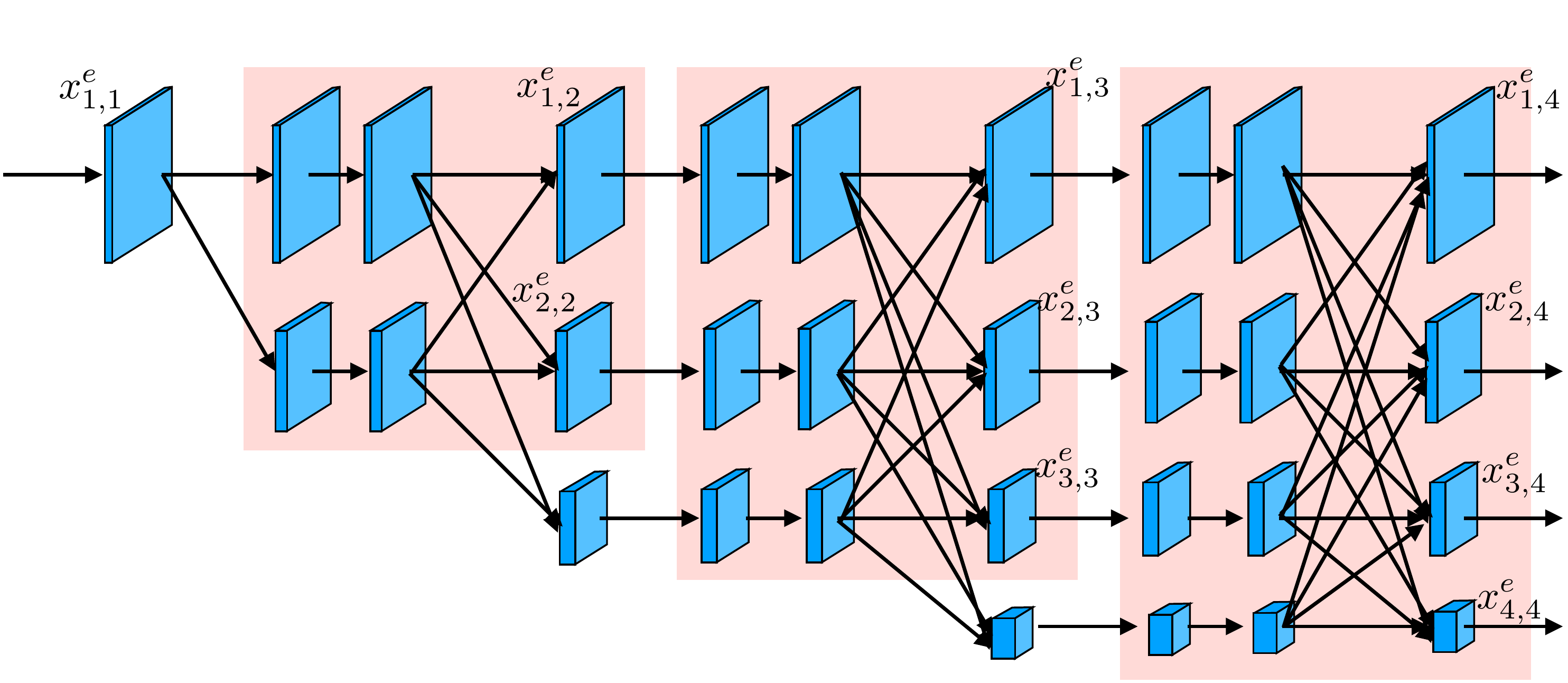}}&
\bmvaHangBox{\includegraphics[height=2.4cm]{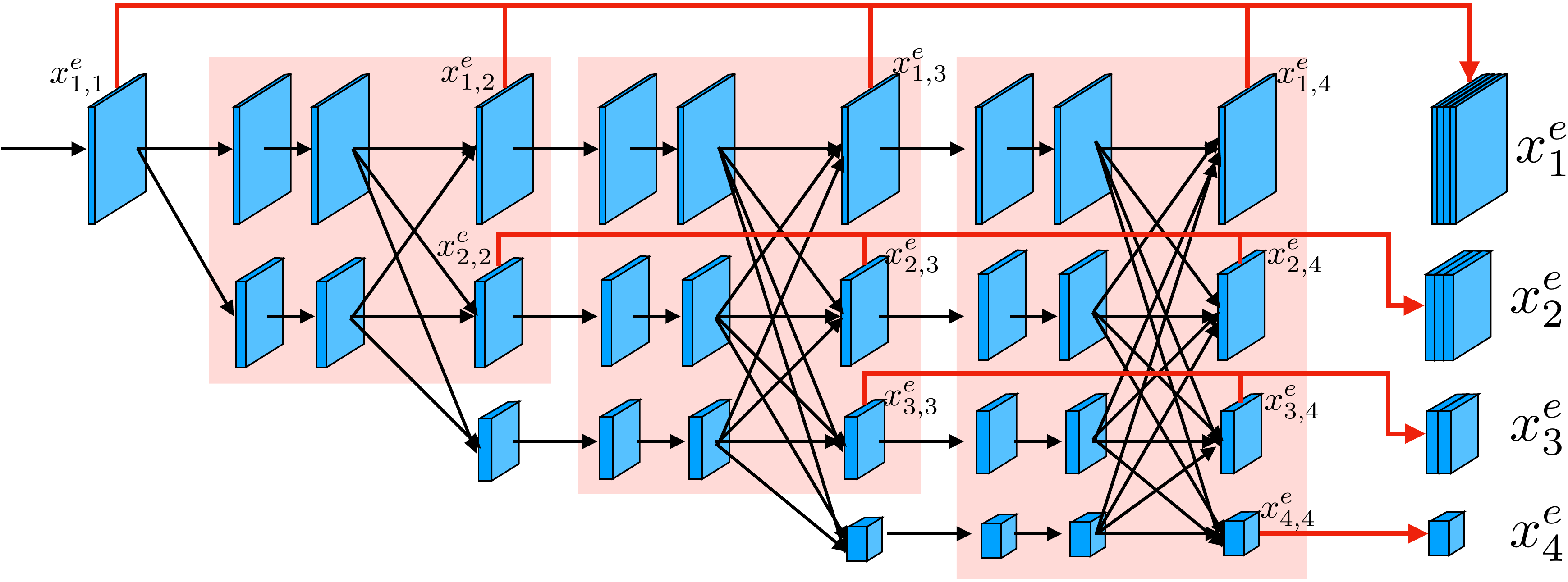}}\\
(a) HRNet &(b) DIFFNet
\end{tabular}
\caption{(a) Original HRNet and (b) DIFFNet architecture with internal feature fusion.}
\label{fig:hrnetvdiffnet}
\end{figure}

High-Resolution Network (HRNet)~\cite{hrnet2020} maintains high resolution representations by the feature extraction process, with two key design characteristics: 
multiple streams with every feature map in the stream having the same resolution, and multiple stages having different resolution exchanging information in each stage. 
HRNet is illustrated in Figure~\ref{fig:hrnetvdiffnet}(a) showing each stage as a red box and each stream as a row. Let $x_{r,s}^e$ denote the feature map from an HRNet encoder node located in the $r$th sub-stream and at the $s$th stage. The resolution of sub-stream $r$ is $\frac{1}{2^{r-1}}$ of the resolution of the first stream. As $r$ increments, the number of channels in the feature maps doubles.  

\begin{figure*}[ht]
\centering
\includegraphics[width=\linewidth]{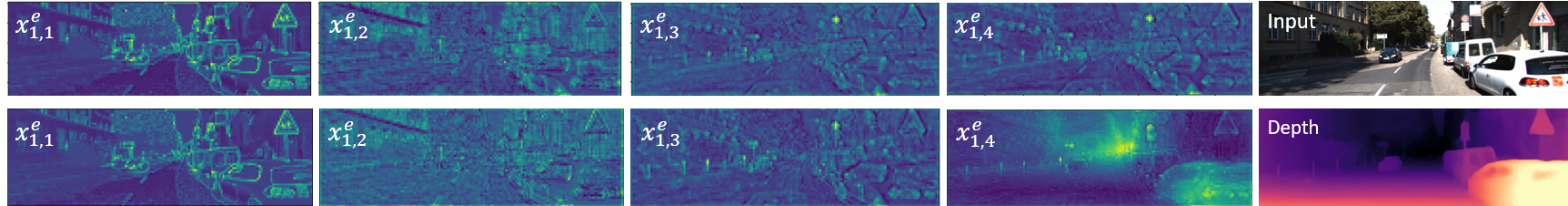}
\caption{Visualisation of intermediate feature maps. We show four intermediate feature maps from stream $r=1$ and stages $s=1,2,3,4$ in the HRNet~\cite{hrnet2020} (top) and DIFFNet (bottom) encoders. 
The final column shows the RGB input and DIFFNet predicted depth map.}
\label{fig:fuse_viz}
\end{figure*}

When we use an HRNet as the encoder for our depth network, we observe significant improvements over other approaches that use ResNet as the encoder.
An HRNet has four streams and four stages, and outputs five feature maps at different scales from the final stage, 
$x_{0,0}^e$ and $x_{r,4}^e, r= 1,2,3,4$. Information from features in previous stages is ignored. We augment this module with internal feature fusion to further exploit the potential of the HRNet architecture:

\noindent\textbf{Multi-stage Internal Feature Fusion} Based on the relationship between feature resolution and spatial information, we assume that feature maps with more channels contain more semantic information and vice versa. To get a semantically-rich intermediate feature map without changing the scale we could increase the number of convolution kernels. However, this would dramatically increase the computational complexity. For example, given a $C_{in}$ dimensional feature and a kernel with a size $3 \times 3$ to output a $C_{out}$ dimensional feature, the number of trainable parameters is $C_{in} \times C_{out} \times  3 \times 3$. If we need double $C_{out}$, the number of parameters also doubles. HRNet contains a multi-stage convolution strategy (Figure~\ref{fig:hrnetvdiffnet}a), and so increasing the convolution kernels leads to a large increase in parameters.
% impossibility for deployment on real-time tasks. 
However, DIFFNet forces feature maps from different stages to contain different semantic information but fuses outputs from all intermediate stages using a concatenation strategy before decoding.
% treats the feature maps from the same stream but different stages equally, which means those to force feature maps from different stage to contain different semantic information but fuse outputs from all intermediate stages using a concatenation strategy before decoding.
Without additional parameters, this strategy is capable of extracting richer feature maps -- see column four in Figure~\ref{fig:fuse_viz}, which shows a smaller semantic gap between DIFFNet encoded features and decoded outputs. 

The stack of feature maps for stream $r$ is computed as:
\begin{equation}\label{eq:stack}
   x_r^e = [x_{r,s}^e],    \qquad     s = r, \cdots,4
\end{equation}
where $[\cdot]$ is the concatenation layer. 
The modified architecture is illustrated in Figure \ref{fig:hrnetvdiffnet}b in which the red arrows denote a concatenation of feature maps. The advantages of giving low level feature maps more semantic information (stacking multi-stage features) is explored in Section~\ref{sec:ablation}.

\subsection{Attention-based Depth Decoder}\label{sec:decoder}
Our decoder is based on a U-Net architecture with further inspiration taken from~\cite{lyu2020hr,hu2018squeeze,woo2018cbam}. Specifically, we introduce an attention mechanism to process the skip-connections from the encoder. An illustration of the decoder can be seen in Figure~\ref{fig:overview} with an outline of each decoder node,  $D_i$, shown bottom right. Let $x_i^d$ denote the output of decoder node $D_i$,\\ calculated as: 
\begin{equation}\label{decoder}
  \left\{
   \begin{array}{lr}
   x_4^d = \mathcal{D}(\sigma([\mu(x_4^e), x_3^e])),& \\
   x_i^d = \mathcal{D}(\sigma([\mu(x_{i+1}^d), x_{i-1}^e])), \qquad i = 1 ,2,3\\
   x_0^d = \mathcal{D}(\sigma(\mu(x_1^d))) &
   \end{array}
  \right.
\end{equation}
where $\mu(\cdot)$ is an upsampling operator, $\sigma(\cdot)$ is an attention module, $[\cdot]$ is concatenation layer and $\mathcal{D}(\cdot)$ is a $3 \times 3$ convolution layer.

\noindent\textbf{Attention Module}. We explore three strategies for incorporating attention into the decoder: channel-wise attention, spatial attention and channel-spatial attention. Given a feature map $\mathcal{F}\in \mathbb{R}^{ C \times H \times W }$, the attention aggregated maps $\mathcal{F}_{c,s,cs}^{'}\in \mathbb{R}^{ C \times H \times W }$ are computed  as:
\begin{equation}\label{eq:attention}
   \begin{array}{lr}
   \mathcal{F}_c^{'} = M_c(\mathcal{F}) \bigotimes \mathcal{F}, \\
   \mathcal{F}_s^{'} = M_s(\mathcal{F}) \bigotimes \mathcal{F}, \\
   \mathcal{F}_{cs}^{'} = M_s(\mathcal{F}_c^{'}) \bigotimes \mathcal{F}_c^{'}.
   \end{array}
\end{equation}
where $M_c(\cdot)$ and $M_s(\cdot)$ are attention map generators which output a 1D channel attention map $m_c\in \mathbb{R}^{ C \times 1 \times 1 }$ and a 2D spatial attention map $m_s\in \mathbb{R}^{ 1 \times H \times W }$ respectively, and $\bigotimes$ denotes element-wise multiplication. During multiplication, the attention values are copied accordingly with channel attention values being broadcast along the spatial dimension, and vice versa (see~\cite{woo2018cbam} for details). We compare these three attention strategies in Section~\ref{sec:ablation} and identify that channel-wise attention gives the best performance. 

% 
%-------------------------------------------------------------------------
% 
\section{Experiments}
In this section, we validate that our proposed network can output semantically-rich and spatially-precise depth maps, and our contributions improve the representation learning ability of HRNet while outperforming other published methods on the KITTI benchmark~\cite{kitti}. Furthermore, we analyse the characteristics of the more challenging scenes from the test partition of the KITTI dataset, and publish identifying information for the high error images. 
\subsection{Dataset}
{\bf KITTI}~\cite{kitti} is a dataset that contains stereo images and corresponding 3D laser scans of outdoor scenes captured by imaging equipment mounted on a moving vehicle~\cite{kingma2015}. The RGB images have a resolution of $\approx 1241 \times 376$ and the corresponding depth maps are sparse with a large amount of missing data. For training, we adopt the dataset split proposed by~\cite{Eigen2014}. After removing the static frames by a pre-processing step suggested by~\cite{zhou_sfmlearner}, this results in 39,810 monocular frame triplets for training and 4,424 frame triplets for validation. To simplify the training process, the camera intrinsic matrices are assumed identical for all the frames in different scenes. To obtain this ``universal'' intrinsic matrix, we offset the principal point of the camera to the image centre and reset the focal length as the average of all the focal lengths in KITTI. This assumption is only valid when the capturing cameras are similar.
\subsection{Implementation Details}
Our models are trained and tested on a single NVidia RTX 6000 GPU using Pytorch~\cite{paszke2019pytorch}. A depth network and a pose network are trained for 20 epochs using the Adam optimizer~\cite{kingma2015} with the default betas $0.9$ and $0.999$. They were trained with a batch size of 16 and an input and output resolution of \( 640 \times 192 \).
We set the initial learning rate as \( 10^{-4} \) for the first 14 epochs and then \( 10^{-5} \) for fine-tuning the remainder. In the objective function $\ell_{final}$ (Equation~\ref{eq:lossfunction}), we let the SSIM weight $\alpha = 0.85$ and the edge-aware smoothness weight $\beta = 1\times 10^{-3}$.

\textbf{Depth Network}. We implement our proposed DIFFNet as described in Section~\ref{sec:diffnet} as our backbone. We use HRNet pre-trained only on ImageNet~\cite{imagenet} to initialize DIFFNet (the effect of pre-training is shown in Table~\ref{tab:ablation}). At training, losses from four scaled depth maps are averaged. When testing, only the maximum resolution depth map is output by the model.

\textbf{Pose Network}. We implement the architecture proposed in~\cite{monodepth2} for pose estimation, which is built on ResNet-18. The pose network takes the two adjacent frames as input and outputs the relative pose which is parameterized with a 6-DOF vector. We experimented with replacing the pose encoder with HRNet, but did not achieve the same performance gains that we observe with the depth network.
% 
%------------------------------------------------------------------------
\subsection{Evaluation on KITTI}
Using metrics described in~\cite{Eigen2014}, we evaluate the performance of DIFFNet on KITTI. The quantitative results are summarized in Table~\ref{tab:results}. Our method outperforms state-of-the-art approaches in terms of Absolute Relative Error and RMSE. When trained on the stereo examples in KITTI, our method achieves best results on all metrics. Given a higher image resolution of $1024 \times 320$, the accuracy of DIFFNet further increases while continuing to outperform competing methods (see in supplementary material for more details).
In Figure~\ref{fig:comparison} we illustrate the qualitative performance of DIFFNet against PackNet~\cite{guizilini2020}, HR-depth~\cite{lyu2020hr} and Monodepth2~\cite{monodepth2}. 
DIFFNet outperforms all self-supervised approaches and even those which use semantic labels as an external supervision resource. We draw attention to the second row that shows our method, where we have used a dashed outline to illustrate the benefits of our semantic backbone when compared with other methods.
We achieve greater detail in a number of roadside items, while holding the advantage of fewer trainable parameters than the other techniques (see Table~\ref{tab:comparison_hard}).
\begin{table*}[t]
\caption{Results on KITTI Benchmark using the Eigen split grouped by training methodology. M: trained on monocular videos, MS: trained on binocular videos. Se: trained with semantic labels. The best scores are {\bf bold} and the second are \underline{underlined}.  }
\label{tab:results}
\centering
\setlength
\tabcolsep{4pt}{
\footnotesize
\begin{tabular}{|l |c|c||c c c c| c c c|}
\hline
%Method & Abs Rel$\downarrow$& Sq Rel$\downarrow$ &  RMSE$\downarrow$ & RMSE log$\downarrow$ %
%& $\delta_{1} < 1.25 \uparrow$ & $\delta_{2} < 1.25^2 \uparrow$ & $\delta_{3} < 1.25^3 \uparrow$  \\ 
\multirow{2}*{Method} & \multirow{2}*{Train} &\multirow{2}*{WxH} &\multicolumn{4}{|c|}{lower is better} & \multicolumn{3}{|c|}{higher is better}\\
~ & ~ & ~ &Abs Rel& Sq Rel&  RMSE & RMSE log & $\delta_{1}$ & $\delta_{2}$ & $\delta_{3} $  \\
\hline
SfMlearner~\cite{zhou_sfmlearner}& M & 640x192 & 0.183 & 1.595 & 6.709 & 0.270 & 0.734& 0.902 &0.959 \\
Li~\cite{li2020unsupervised}& M& 416x128  & 0.130 & 0.950 & 5.138 & 0.209 & 0.843 & 0.948 & 0.978\\
Chen~\cite{chen2019towards}& M+Se & 512x256   & 0.118 & 0.905 & 5.096 & 0.211 & 0.839 & 0.945 &0.977\\
Monodepth2~\cite{monodepth2}& M & 640x192   & 0.115 & 0.903 & 4.863 & 0.193 & 0.877 & 0.959 & 0.981\\
SGDepth~\cite{klingner2020self}& M+Se & 640x192  & 0.113 & 0.835 & 4.693 & 0.191 & 0.879 & 0.961 & 0.981\\
SAFENet~\cite{choi2020safenet}& M+Se & 640x192 & 0.112 & 0.788 & 4.582 & 0.187 & 0.878 & 0.963 & \bf{0.983} \\
VC-Depth~\cite{zhou2020constant}& M & 640x192   & 0.112 & 0.816 & 4.715 & 0.190 & 0.880 & 0.960 & 0.982\\
PackNet~\cite{guizilini2020}& M & 640x192 & 0.111 & \underline{0.785} & 4.601 & 0.189 & 0.878 & 0.960 & 0.982\\
Mono-Uncertainty\cite{poggi2020uncertainty}& M & 640x192 & 0.111  & 0.863 & 4.756 & 0.188 & 0.881 & 0.961 & 0.982\\
Fang~\cite{fang2020towards}& M & 640x192 & 0.111 & - & 4.660  & 0.186 & 0.884 & 0.962 & 0.982 \\
HR-depth~\cite{lyu2020hr}& M & 640x192   & 0.109 & 0.792 & \underline{4.632} & 0.185& 0.884 & 0.962 & \bf{0.983}\\
%UnRectDepth~\cite{kumar2020unrectdepthnet}& M & 640x192 & 0.107 & \bf{0.721} & \underline{4.564} & {\bf0.178} &\bf{0.894} & {\bf{0.971}} & {\bf 0.986} \\
Johnston~\cite{johnston2020self}& M & 640x192 & \underline{0.106} & 0.861 & 4.699 & \underline{0.185} & \underline{0.889} & \underline{0.962} & \underline{0.982} \\
%SynDistNet~\cite{kumar2021syndistnet}& M+Se & 640x192 & 0.109 & \bf{0.718} & \underline{4.516} & 0.180 & \bf{0.896} & \bf{0.973} & \bf{0.986} \\
%Guizilini~\cite{guizilini2020semantically} & 0.102 & 0.698 & 4.381 & 0.178 & 0.896 & 0.964 & 0.984 \\
%Manydepth*~\cite{watson2021temporal} & 0.098 & 0.770 & 4.459 & 0.176 & 0.900 & 0.965 &0.983 \\
%CoMoDA~\cite{kuznietsov2021comoda}& M* & 640x192  & 0.103 & 0.862 & 4.594 & 0.183 & 0.899 & 0.961 & 0.981\\
\hline

\bf{DIFFNet} & M & 640x192 & \bf{0.102}  &   \bf{0.764}  &   \bf{4.483}  &   \bf{0.180}  &   \bf{0.896}  &   \bf{0.965}  &   \bf{0.983}  \\
%conv1x1 (Ours) & M & 640x192    &   \bf{0.103}  &   0.744  &   \bf{4.474}  &\underline{0.179}    &   0.893  &   0.964  &   \underline{0.983}    \\									
%hrnet-18+VC& M & 640x192 & 0.107 & 0.778 & 4.588 & 0.185 & 0.889 & 0.963 & 0.982\\
% HRNet-32 (31M)& M & 640x192 &\underline{0.104} & 0.772 & \underline{4.498} & 0.180 & \underline{0.894} & 0.965 & 0.983\\
% HRNet-48 (68M)& M & 640x192 & \underline{0.104} & 0.878 & 4.678 & 0.184 & {\bf 0.895} & 0.963 & 0.981\\
% HRNet-64 (119M)& M & 640x192 & {\bf 0.103} & 0.844 & 4.681 &  0.184 & {\bf0.895} & 0.963 & 0.982\\
\hline
\hline
Monodepth2~\cite{monodepth2}& MS & 640x192   & \underline{0.106} & 0.818 & 4.750 & 0.196 & 0.874 & 0.957 & 0.979\\
HR-depth~\cite{lyu2020hr}& MS & 640x192   & 0.107 & \underline{0.785} & 4.612 & \underline{0.185} & \underline{0.887} & \underline{0.962} & \underline{0.982}\\
Fang~\cite{fang2020towards}& MS & 640x192 & \bf{0.101} & - & \underline{4.512}  & 0.188 & 0.881 & 0.961 & 0.981 \\
\hline
\bf{DIFFNet} & MS & 640x192 &   \bf{0.101}  &   \bf{0.749}  &   \bf{4.445}  &   \bf{0.179}  &   \bf{0.898}  &   \bf{0.965}  &   \bf{0.983}\\
%conv1x1 (Ours) & MS & 640x192 &   \bf{0.101}  &   \bf{0.743}  &   \bf{4.475}  &   \bf{0.179}  &   \bf{0.896}  &   \bf{0.964}  &   \bf{0.983}  \\
%hrnet-32 & MS & 640x192&   0.098  &   0.724  &   4.407  &   0.178  &   0.903  &   0.965  &   0.983  \\						
%hrnet-48 & MS & 640x192&   0.097  &   0.707  &   4.379  &   0.177  &   0.904  &   0.966  &   0.983  \\
%hrnet-64 & MS &  640x192& 0.098  &   0.727  &   4.415  &   0.178  &   0.904  &   0.965  &   0.982  \\
\hline
\hline
Monodepth2~\cite{monodepth2}& M & 1024x320 & 0.115 & 0.882 & 4.701 & 0.190 & 0.879 & 0.961 & 0.982\\
Fang~\cite{fang2020towards}& M & 1024x320 & 0.109 & - & 4.581  & 0.185 & 0.890 & 0.964 & \underline{0.983} \\
PackNet~\cite{guizilini2020}& M & 1280x384 & 0.107 & 0.802 & 4.538 & 0.186 & 0.889 & 0.962 & 0.981\\
SGDepth~\cite{klingner2020self}& M+Se & 1280x384  & 0.107 & 0.768 & 4.468 & 0.186 & 0.891 & 0.963 & 0.982\\
SAFENet~\cite{choi2020safenet}& M+Se & 1024x320 & 0.106 & 0.743 & 4.489 & 0.181 & 0.884 & 0.965 & \bf{0.984} \\
HR-depth~\cite{lyu2020hr}& M & 1024x320 & 0.106 & 0.755 & 4.472 & 0.181 & 0.892 & \underline{0.966} & \bf{0.984}\\
Feat-Depth~\cite{shu2020feature}& M & 1024x320   & 0.104& \underline{0.729} & 4.481&0.179 & 0.893 & 0.965 & \bf{0.984}\\
%UnRectDepth~\cite{kumar2020unrectdepthnet}& M & 1024x320 & 0.103 & \bf{0.705}& 4.386  & \bf{0.164} & 0.897 & \bf{0.980} & \bf{0.989} \\
Guizilini~\cite{guizilini2020semantically}& M+Se & 1280x384 & \underline{0.100} & 0.761 & \bf{4.270} & \underline{0.175} & \underline{0.902} & 0.965 & 0.982 \\
%Feat-Depth~\cite{shu2020feature}& MS & 1024x320   & 0.099 & 0.697 & 4.427 & 0.184 & 0.889 & 0.963 & 0.982\\
%Zhu~\cite{Zhu_2020_CVPR}& S+Se & 1024x320 & 0.097& 0.675 & 4.350& 0.180&0.890&0.964&0.983\\
\hline
%hrnet18& M & 1024x320 & 0.098 & 0.766 & 4.395 & 0.176 & 0.906 & 0.967 & 0.983\\
%conv1x1 (Ours)& M & 1024x320 & \bf{0.097} & \bf{0.705} & \underline{4.343} & \underline{0.175} & \bf{0.905} & \underline{0.966} & \underline{0.984}\\
\bf{DIFFNet}& M & 1024x320 &   \bf{0.097}  &  \bf{ 0.722}  &   \underline{4.345}  & \bf{0.174}    &   \bf{0.907}  &   \bf{0.967}  &   \bf{0.984}  \\

\hline
\end{tabular}}
\end{table*}
\begin{figure*}[ht]
\centering
\includegraphics[width=\linewidth]{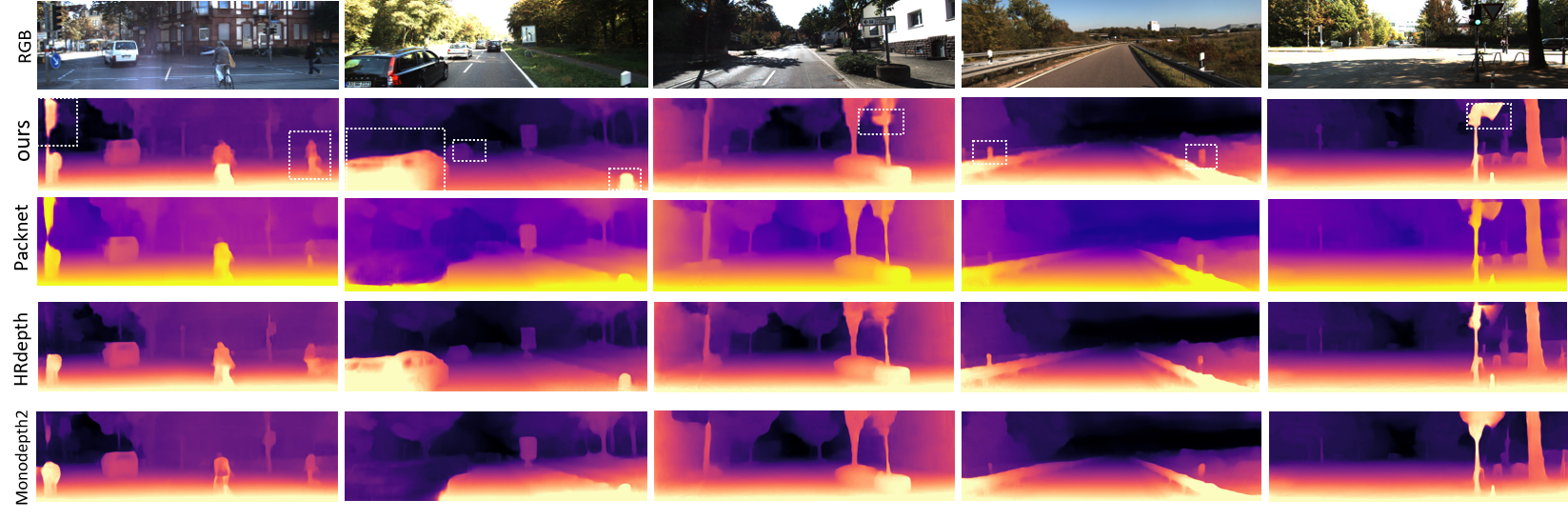}
\caption{Visualisation of depth estimation results. The top row contains the input images. 
    The second row shows the result from DIFFNet, and the remaining rows are from other contemporary methods.
    Note the improvement in detail for many roadside items, that our semantic backbone provides.
    Hotter colours indicate closer objects.}
\label{fig:comparison}
\end{figure*}
% 
% 
%------------------------------------------------------------------------
% 
\subsection{Ablation Study}\label{sec:ablation}
\begin{figure*}
 \centering
 \includegraphics[ width=\linewidth]{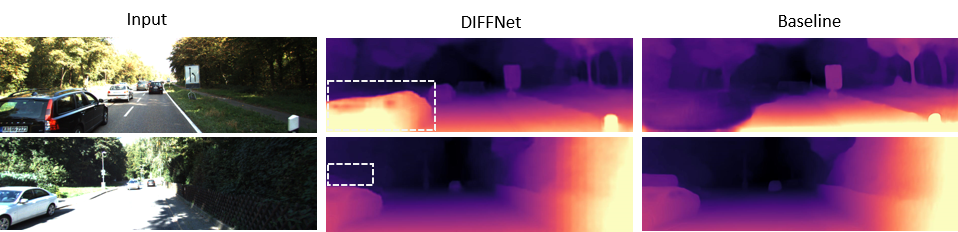}
 \caption{Visualisation of the ablation study. Row one shows that with more semantic information fed into depth decoder, the predicted depth map will more precise. Row two shows that DIFFNet produces a depth map with fewer artefacts than the baseline method.}
 \label{fig:viausal_ablation}
 \end{figure*}

To validate the performance improvements that our contributions provide, we conduct an ablative analysis.
We establish a baseline by replacing the original ResNet-based depth encoder in Monodepth2~\cite{monodepth2} with HRNet-18. Table~\ref{tab:ablation} shows the results of the analysis,
with the progressive addition of pre-training the encoder on ImageNet, multi-stage fusion (MF), channel-wise attention (CA) and space-wise attention (SA). 
The largest performance gain is achieved by pre-training the encoder rather than training from scratch.
We observe that channel-wise attention yields increased accuracy compared with spatial attention. 
Furthermore, feature fusion improves baseline performance for all attention configurations with the exception of channel-spatial. 
A qualitative comparison of DIFFNet and the baseline model is shown in Figure~\ref{fig:viausal_ablation}.
% 
% REMOVE ABLATION FIGURE FOR NOW
% \begin{figure*}
% \centering
 % [trim=left bottom right top, clip]
% \includegraphics[trim=0 15 20 0, clip, width=\linewidth]{images/ablation4.png}
% \caption{Visualisation of ablation studies. HRNet-64: replacing the encoder of %\cite{monodepth2} with it. (a) shows that with more semantic information fed into depth %decoder, the predicted depth map will more precise even using a small model. (b) shows %ours produces a depth map with fewer artefacts.}
% \label{fig:viausal_ablation}
% \end{figure*}
% 
\begin{table*}[ht]
\caption{\bf{Ablation Studies.} MF: Multi-stage Fusion. CA: Channel-wise Attention. SA: Space-wise Attention. Red check marks identify our final system.}
\label{tab:ablation}
\centering
\setlength
\tabcolsep{4.2pt}{
\footnotesize
\begin{tabular}{|l|c |c | c c||c c c c | c c c|}
\hline
\multirow{2}*{Method} & \multirow{2}*{Pre-train} & Encoder  &\multicolumn{2}{|c||}{Decoder} &\multicolumn{4}{|c|}{The lower the better} & \multicolumn{3}{|c|}{The higher the better}\\
~ & ~ & MF & CA & SA &Abs Rel& Sq Rel&  RMSE & RMSE log & $\delta_{1}$ & $\delta_{2}$ & $\delta_{3} $  \\
\hline
 %HRNet-32&  & &  & 0.104 & 0.772 & 4.498 & 0.180 & 0.894 & 0.965 & 0.983    \\
 %HRNet-48&  &  &  & 0.104 & 0.878 & 4.678 & 0.184 & 0.895 & 0.963 & 0.981   \\\hline
 
\multirow{2}*{Baseline}& & & & &   0.124  &   0.990  &   5.158  &   0.202  &   0.858  &   0.952  &   0.974  \\
~&\checkmark&&&&   0.108  &   0.799  &   4.609  &   0.186  &   0.888  &   0.963  &   0.982\\\hline
%~& & \checkmark& &    0.105  &   0.793  &   4.551  &   0.182  &   0.892  &   0.964  &   0.983    \\
%~& & &\checkmark &   0.106  &   0.886  &   4.715  &   0.185  &   0.892  &   0.963  &   0.982    \\
%~& & \checkmark &\checkmark &   0.104  &   0.799  &   4.596  &   0.183  &   0.890  &   0.963  &   0.983  \\\hline
\multirow{5}*{DIFFNet}& &\checkmark &\checkmark&&  0.119  &   0.937  &   4.905  &   0.198  &   0.867  &   0.955  &   0.979  \\
~ &\checkmark & \checkmark & &   &   0.105  &   0.817  &   4.593  &   0.183  &   0.893  &   0.964  &   0.982  \\
%~ & \color{red}\checkmark & \color{red}\checkmark &  & \bf{0.103}  &   0.744  &   4.474  &   \bf{0.179}  &   \bf{0.893}  &   0.964  &   0.983    \\
~& \color{red}\checkmark & \color{red}\checkmark & \color{red}\checkmark &  & \bf{0.102}  &   \bf{0.764}  &   \bf{4.483}  &   \bf{0.180}  &   \bf{0.896}  &   \bf{0.965}  &   \bf{0.983}   \\
~& \checkmark & \checkmark & & \checkmark &   0.107  &   0.822  &   4.637  &   0.183  &   0.890  &   0.963  &   0.983  \\
~ & \checkmark& \checkmark & \checkmark & \checkmark &   0.103  &   0.769  &   4.530  &   0.180  &   0.892  &   0.964  &   0.983  \\
%  ~ & HRNet-64  & \checkmark &  & 0.103 & 0.844 & 4.681 &  0.184 & 0.895 & 0.963 & 0.982\\
\hline
\end{tabular}}
\end{table*}
% 
%------------------------------------------------------------------------
% 
\subsection{Extended Evaluation}
% 
%\begin{figure*}[ht]
%\centering
%\includegraphics[width=\linewidth]{images/intersection.png}
%\caption{We create a hard test set of 23 images that is the union set of %the ten highest error images from recent well known works (Table %\ref{tab:comparison_hard}). Here we illustrate the \emph{intersection} %set of 3 images and our qualitative results. }
%\label{fig:intersection}
%\end{figure*}
% 
Table~\ref{tab:results} reveals the relative performance gap between contemporary methods on KITTI is diminishing. 
From empirical testing, we observe that the 10 images that give the highest error from each of these methods represents $\approx 1.4\%$ of the KITTI test set, but contributes $> 3\%$ of error when evaluating. 
Hence, error is not uniformly distributed throughout the test set, but certain images are more challenging than others.
A model's performance on its own top 10 hard cases is a key factor in measuring its robustness and stability. 
For a fair comparison, we propose that the difficult cases from competing methods form a single challenge set. 
It is our hope that future authors will accept this strategy when they evaluate their models and compare against others.

In our case, we create a challenging test set that is the union of the 10 images with highest error from the four approaches shown in Table \ref{tab:comparison_hard}, 
including a baseline method discussed in Section \ref{sec:ablation}. 
The union set comprises 23 images\footnote{Indices in KITTI benchmark: 58, \textbf{68}, \textbf{73}, \textbf{106}, 164, 173, 183, \textbf{260}, \textbf{330}, \textcolor{red}{\textbf{374}}, 377, 385, 386, \textbf{388}, \textcolor{red}{\textbf{394}}, \textcolor{red}{\textbf{395}}, 477, 504, 518, 548, \textbf{549}, 559, 683. Those from ours are \textbf{bold} and common hard cases are \textcolor{red}{red}. The corresponding images are shown in the supplementary material. }, and 3 images are common to all sets.
In Table \ref{tab:comparison_hard} it is clear that our method performs competitively under this most difficult test, resulting in the lowest Absolute Relative Error. 
%
%We also show a qualitative result on the intersection set of 3 images in Figure \ref{fig:intersection}. 
We can hypothesise these are the most challenging images due to the large regions of foliage in combination with difficult lighting. 
\begin{table*}[ht]
\caption{Quantitative results on the challenging KITTI examples.}
%The baseline method is described in our ablation study, discussed in Section \ref{sec:ablation}.}
\label{tab:comparison_hard}
\centering
\setlength
\tabcolsep{4.2pt}{
\footnotesize
\begin{tabular}{|l|c| c | c c c c | c c c|}
\hline
\multirow{2}*{Method} & \multirow{2}*{Parameters} & Run-time &\multicolumn{4}{|c|}{lower is better} & \multicolumn{3}{|c|}{higher is better}\\
 & & FPS &Abs Rel& Sq Rel&  RMSE & RMSE log & $\delta_{1}$ & $\delta_{2}$ & $\delta_{3} $  \\
\hline
Monodepth2~\cite{monodepth2}& 14.84M &  99 & 0.213  &   2.197  &   6.468  &   0.295  &   0.741  &   0.906  &   0.950  \\
%packnet~\cite{guizilini2020}&   0.202 &   0.764  &   4.583  &   0.185  &   0.888  &   0.963  &   0.983  \\
%Baseline & 11.31M &   0.203  &   1.613  &   5.779  &   0.299  &   0.729  &   0.903  &   0.956  \\
HR-Depth~\cite{lyu2020hr} & 14.62M &  116  & 0.205  &   \bf{1.591}  &   \bf{5.726}  &   
0.282  &   0.738  &   0.902  &   0.957  \\
\hline
\textbf{DIFFNet}&   10.8M &  87  & \bf{0.197}  &   1.803  &   5.988  &   \bf{0.282}  &   \bf{0.763}  &   \bf{0.912}  &   \bf{0.957}  \\
\hline
\end{tabular}}
\end{table*}
% 
% \begin{figure*}
%  \centering
%  \includegraphics[ width=\linewidth]{images/ablation4.png}
%  \caption{\textcolor{red}{Visualisation of ablation study result. Row one shows that with more semantic information fed into depth decoder, the predicted depth map will more precise. Row two shows that DIFFNet produces a depth map with fewer artefacts than the baseline method.}}
%  \label{fig:viausal_ablation}
%  \end{figure*}
%------------------------------------------------------------------------
\section{Conclusion}
In this work, we have proposed DIFFNet for self-supervised monocular depth estimation. Based on HRNet, which is designed for other computer vision tasks, we adopt it and improve it with two simple but effective strategies. Specifically, we incorporate multiple resolution feature fusion and a channel attention mechanism. With fewer parameters to learn, DIFFNet outperforms other state-of-the-art self-supervised methods, especially when high resolution input is available. 
We have shown that the DIFFNet encoder computes semantically rich feature maps, and our ablation study demonstrates the performance gain from each proposed modification.
Finally, we introduced a creative strategy for evaluating models by investigating difficult test cases, and we invite authors to adopt the same approach going forward. 
% 
%--------------------------------
\\[5pt]
\textbf{Acknowledgement}\\
The research presented in this paper was carried out on the High Performance Computing Cluster supported by the Research and Specialist Computing Support service at the University of East Anglia.

\bibliography{bmvc2021}
\end{document}